\pdfoutput=1

\documentclass[11pt]{article}

\usepackage[]{acl}

\usepackage{times}
\usepackage{latexsym}

\usepackage[T1]{fontenc}

\usepackage[utf8]{inputenc}

\usepackage{microtype}

\usepackage{amsmath}
\usepackage{booktabs}
\usepackage{graphicx}
\usepackage{multirow}
\usepackage{float}

\title{Anaphora Resolution in Dialogue: System Description (CODI-CRAC 2022 Shared Task)}

\author{{\bf Tatiana Anikina} \\  {\bf Natalia Skachkova} \\  DFKI / Saarland Informatics Campus, \\ Saarbr\"ucken, Germany \\ \texttt{tatiana.anikina@dfki.de} \\ \texttt{natalia.skachkova@dfki.de} \And
{\bf Joseph Renner} \\ {\bf Priyansh Trivedi} \\ Inria, \\ Nancy, France \\\texttt{joseph.renner@inria.fr} \\ \texttt{priyansh.trivedi@inria.fr}}

\begin{document}
\maketitle
\begin{abstract}
We describe three models submitted for the CODI-CRAC 2022 shared task. To perform identity anaphora resolution, we test several combinations of the incremental clustering approach based on the Workspace Coreference System (WCS) with other coreference models. The best result is achieved by adding the ``cluster merging'' version of the \emph{coref-hoi} model, which brings up to 10.33\% improvement\footnote{An average improvement over all 4 datasets is 7.95\%.} over vanilla WCS clustering. 
Discourse deixis resolution is implemented as multi-task learning: we combine the learning objective of \emph{coref-hoi} with anaphor type classification. 
We adapt the higher-order resolution model introduced in \citet{joshi-etal-2019-bert} 
for bridging resolution given gold mentions and anaphors. 
\end{abstract}

\section{Introduction}

In this paper we present our systems submitted for the CODI-CRAC 2022 Shared Task (CCST) on Anaphora, Bridging, and Discourse Deixis in Dialogue\footnote{\url{https://codalab.lisn.upsaclay.fr/competitions/614\#learn_the_details}}~\cite{khosla-etal-2022-codi}. The task is a follow-up to the one held last year and described in~\citet{khosla-etal-2021-codi}. As its name suggests, besides identity anaphora this shared task tries to cover other, less-studied, anaphoric phenomena, and offers new multi-genre data that combines several types of annotations in Universal Anaphora\footnote{\url{https://universalanaphora.github.io/UniversalAnaphora/}} format. 

Main focus of the shared task is on dialogue. Dialogue data offers new challenges, like grammatically incorrect utterances, disfluencies, more deictic references,
speaker grounding and long-distance conversation structure~\cite{khosla-etal-2021-codi}. While coreference resolution in text has been very actively studied in the recent years, it is much less researched in dialogue, especially such forms as bridging, or discourse deixis. Descriptions of early systems implemented for the resolution of `standard' and discourse deictic pronouns in dialogue can be found, e.g., in~\citet{byron2002resolving}, \citet{strube-muller-2003-machine}, \citet{muller2008fully}. More approaches (not implemented), together with some useful findings are presented, e.g., in~\citet{rocha-1999-coreference}, \citet{eckert2000dialogue}, and \citet{navarretta-2004-resolving}.

CCST 2021 stirred new interest in coreference resolution in dialogue. The majority of systems submitted for it represent various modifications of either the higher-order coreference resolution model (\emph{coref-hoi}) by~\citet{xu-choi-2020-revealing}, or one of the earlier models by~\citet{joshi-etal-2019-bert} and \citet{lee-etal-2018-higher}. These models were originally trained on the text data, and are span-based - each span gets associated with a score, and anaphor-antecedent pairs are established based on the pairwise scores. Designed for identity anaphora resolution, these models were also adapted for bridging and discourse deixis resolution. Examples of span-based models submitted for CCST 2021 include systems by~\citet{kobayashi-etal-2021-neural}, \citet{renner-etal-2021-end}, \citet{xu-choi-2021-adapted}. Other participants presented different approaches. Thus,  \citet{kim-etal-2021-pipeline} perform identity anaphora and bridging resolution using pointer networks. \citet{anikina-etal-2021-anaphora} cast anaphora resolution as a clustering problem, and discourse deixis resolution - as a Siamese Net based scoring function.

Inspired by the success of the span-based coreference resolution models, we submit three independent systems for CCST 2022. Our system for identity anaphora resolution uses both the Workspace Coreference System by~\citet{anikina-etal-2021-anaphora} and the \emph{coref-hoi} model as described in Section~\ref{sec:anaphora-res}. The model for discourse deixis extends \emph{coref-hoi} with shallow linguistic features and aims at resolving three types of potential anaphors. It is described in Section~\ref{sec:dd-res}. The model for bridging resolution is a modification of the system by~\citet{renner-etal-2021-end}. The approach is explained in Section~\ref{sec:bridging-res}.

\section{Anaphora Resolution}
\label{sec:anaphora-res}

\begin{table*}[h]
    \centering
    \scalebox{0.95}{
    \begin{tabular}{l|l}
        \hline
        \bf{Track} & Resolution of anaphoric identities \\
        \hline
        \bf{Setting} & Predicted mentions \\
        \hline
        \bf{Baseline} & WCS \cite{anikina-etal-2021-anaphora} and \emph{coref-hoi} model \cite{xu-choi-2020-revealing} \\
        \hline
        \bf{Approach} & \shortstack[l]{1) Extract all nominal phrases with SpaCy \\
        2) Run WCS trained on the Shared Task dialogue data \\
        3) Run \emph{coref-hoi} with cluster merging trained on the CoNLL 2012 data\\
        4) Combine the outputs of WCS and \emph{coref-hoi}} \\
        \hline
        \bf{Train data} & \shortstack[l]{ARRAU corpus (Gnome, Trains\_91, Trains\_93, RST\_DTreeBank, Pear\_stories), AMI, \\ Switchboard, Light and Pesuasion, CoNLL 2012 English dataset} \\
        \hline
        \bf{Dev data} & AMI, Light, Persuasion, ARRAU (dev splits) \\
        \hline
    \end{tabular}}
    \caption{Anaphora resolution: approach summary}
    \label{tab:AR-summary}
\end{table*}

For the anaphora resolution track we trained and combined the outputs of the Workspace Coreference System (WCS) and the \emph{coref-hoi} system (see Table \ref{tab:AR-summary}). While working on the shared task we realized that a combination of different models performs better than a single model and we explored various settings to find an optimal solution.

\subsection{Data}
\label{subsec:ar-data}

For training of the WCS system we used the datasets recommended by the shared task organizers. These include the ARRAU corpus (Gnome, Trains\_91, Trains\_93, RST\_DTreeBank, Pear\_stories), AMI, Switchboard, Light and Persuasion data. We used the development sets of AMI, Light and Persuasion for the internal evaluation and comparison of different configurations. We trained our system using the gold mention spans to avoid any mistakes introduced by the mention extraction module and used SpaCy~\cite{spacy} for mention extraction during the test phase.

For training of the \emph{coref-hoi} system, we utilized the CoNLL 2012 English
Shared Task dataset \cite{pradhan-etal-2012-conll} to supplement the datasets listed in the previous paragraph. Note that this CoNLL 2012 data does not include singleton coreference clusters, but the current dialogue shared task datasets do.

\subsection{Model architecture}

\paragraph{WCS}

\begin{figure}[h]
\centering
\includegraphics[width=.8\linewidth]{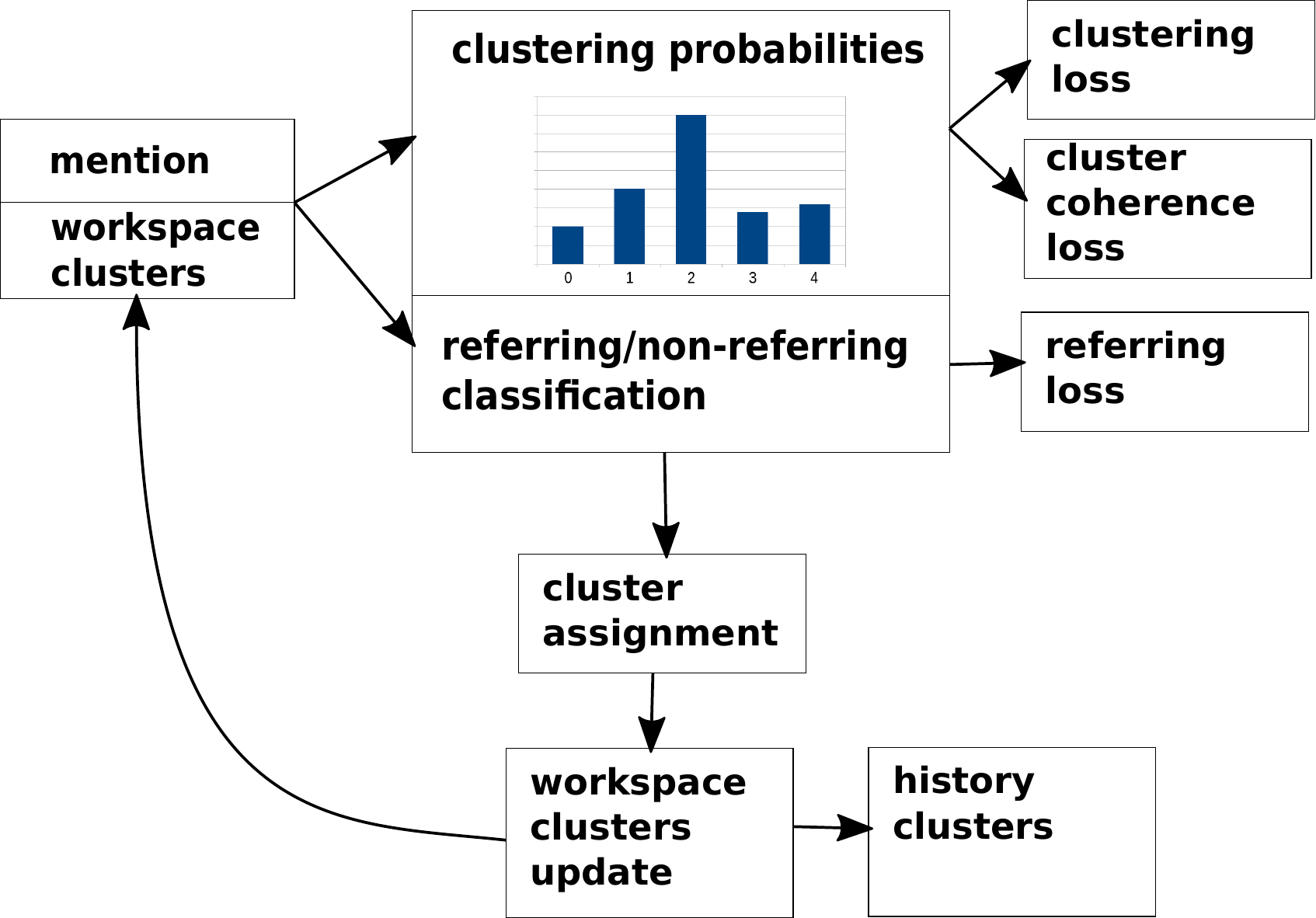}
\caption{Workspace Coreference System Overview}\label{fig:overview}
\end{figure}

Our model is based on the implementation described in~\citet{anikina-etal-2021-anaphora}. It creates coreference clusters incrementally and compares each new mention to the clusters that are available in the workspace. The general flow of the model is presented in Figure \ref{fig:overview}. The model uses separate layers to encode each pair of mentions where one mention represents a workspace cluster and another mention is a candidate that is being clustered. WCS passes the concatenated embeddings of the candidate mention and the cluster member through several feed-forward neural layers with the input and output dimensions shown in Table \ref{tab:wcs-architecture}.

The network also encodes the absolute position of each mention within the document and generates a separate embedding for each speaker. The model combines this information with different word embeddings. For each mention it extracts the head and encodes it with a combination of contextual BERT embeddings~\cite{devlin2018bert}  (\texttt{bert-base-cased}) 
together with GloVe~\cite{pennington2014glove} and Numberbatch~\cite{speer2017conceptnet} embeddings. Unlike~\citet{anikina-etal-2021-anaphora} we do not generate a new random embedding for each unknown word, but take an average embedding based on all words in the GloVe and Numberbatch vocabularies. This gave us slightly better results in the pilot experiments.

\begin{table}[h]
    \centering
    \scalebox{0.8}{
    \begin{tabular}{lccc}
        \toprule
        Encoder & Input dim & Hidden dim & Output dim \\
        \midrule
         BERT head & 2*768 & 900 & 600 \\
         BERT span & 2*768 & 900 & 600 \\ 
         Numberbatch & 2*300 & 600 & 300 \\        
         GloVe head & 2*100 & 600 & 200 \\
         GloVe span & 2*100 & 600 & 200 \\
         BERT masked LM & 2*768 & 600 & 200 \\
        \bottomrule 
    \end{tabular}}
    \caption{Separate encoders are used to represent mention pairs in WCS. Additionally, distance between the mentions, their positions in the document and corresponding speakers are encoded and added to the final representation.}
    \label{tab:wcs-architecture}
\end{table}

In order to represent the spans we take an average of all individual word embeddings based on BERT and GloVe correspondingly. We also experimented with SpanBERT embeddings but did not observe any improvements. 
E.g., when we replaced our span embeddings with SpanBERT and left the rest of the system unchanged we achieved 66.68\% CoNLL F1 score when training and evaluating on the Light dataset. After replacing SpanBERT with standard BERT and simply averaging span embeddings we achieved 67.23\% CoNLL F1 score on the same data. Removing GloVe embeddings and leaving only BERT, SpanBERT and Numberbatch or training on more data samples also did not help. We suspect that since SpanBERT embeddings have high dimensionality (representing span start, span end and span head) they dominate mention representation in WCS and allow some vague semantic matches. E.g., with SpanBERT we generated clusters that included mentions like \textit{`war'} and \textit{`peace'} or \textit{`the jamaica tourist board'} and \textit{`jamaican'}. Training for more epochs or adjusting hyperparameters might help to improve clustering but the configurations that we tested have not shown an improvement. 


The WCS system combines three cross-entropy losses that are added in each forward pass. The main clustering loss compares the true cluster probabilities vs. the computed ones. The true probabilities are computed with respect to the mentions that are currently in the workspace. For each mention the probability of being in that cluster is defined as the ratio of mentions that are in the same gold cluster and the current cluster over all mentions in that cluster. The coherence loss computes the difference between the gold cluster assignments and the system assignments. Basically, we create two matrices that align mentions to each other and check the overlap between these matrices in the gold annotations vs. the generated outputs (the matrix has ones if two mentions belong to the same cluster and zeros otherwise). The referring loss is used for the referring expression classification which is a binary classification task. It is needed since not all mention spans extracted by SpaCy are valid referring expressions.

After computing clustering probabilities for each mention and clusters in the workspace we apply softmax and select the cluster with the highest probability. After that the workspace is updated and some clusters are moved to the history if they have not been updated for more than 100 steps. After the initial clustering we apply some post-processing as explained in~\citet{anikina-etal-2021-anaphora}.

We have also evaluated WCS in combination with a Crosslingual Coreference System (CCS) based on AllenNLP and SpaCy pipelines\footnote{\url{https://pypi.org/project/crosslingual-coreference/}}. We noticed that WCS performs quite well on identifying singletons and clusters with personal pronouns but has more difficulties with other nominal phrases. Hence, in one of the experiments we combined the output of the CCS model trained on OntoNotes 
that uses MiniLM~\cite{DBLP:conf/nips/WangW0B0020} for mention representation with the outputs of WCS trained on the shared task data. 
Among the clusters generated with CCS we selected only those that do not contain any personal pronouns and from WCS we took singletons and clusters with pronouns.

We also experimented with some compatibility checks. E.g., we checked whether the first and second mentions in the cluster have the same number and we removed the first mention from the WCS cluster if the embedding similarity between the first pronoun and the first noun in that cluster was too low (compared to the cosine similarity between the first pronoun and other mentions in the cluster). 
E.g., mentions such as \textit{`a presenter'} and \textit{`I'} could belong to the same cluster with pronouns but mentions like \textit{`table'} and \textit{`I'} should not. We run WCS with these modifications on the shared task test set and report our results in Table ~\ref{tab:wcs-combinations}. The final version that was submitted to the leaderboard combines WCS outputs with the \emph{coref-hoi} system as described in the next section. 

\begin{table}[]
    \centering
    \scalebox{0.8}{
    \begin{tabular}{lcccc}
        \toprule
        Setting & Light & AMI & Persuasion & Swbd. \\
        \midrule
         Vanilla WCS & 65.96 & 46.04 & 59.54 & 50.63 \\
         WCS + CCS & 67.27 & 46.68  & 63.46 & 53.92  \\
         WCS + CCS + filter & 67.46 & 46.70 & 63.51 & 54.07 \\
         WCS + coref-hoi & 72.06 & 51.41  & 69.87 & 60.61  \\         
        \bottomrule 
    \end{tabular}}
    \caption{Evaluation of WCS in combination with other coreference systems on the shared task test set. Filter in the third row refers to the incompatibility check}
    \label{tab:wcs-combinations}
\end{table}

\paragraph{Coref-HOI Combination}
We trained a ``cluster merging'' variant of the \emph{coref-hoi} model. As this model was developed using the data from 2012 CoNLL dataset, which does not include singleton clusters, the model does not output singleton predictions off the shelf (one could potentially use the scores for the ``dummy'' antecedent as a proxy, but this could be noisy as the model is not trained to differentiate singleton clusters from simple mentions that are not part of any cluster). 

\begin{table*}[]
    \centering
    \scalebox{0.8}{
    \begin{tabular}{|l|cc|cc|cc|cc|}
        \toprule
        Setting &  Light & Light NS  & AMI & AMI NS  & Persuasion & Persuasion NS  & ARRAU & ARRAU NS \\
        \midrule
         WCS & 65.39  & 61.48 & 43.33 & 35.85 & 61.23 & 56.55 & 45.02 & 32.93 \\
         coref-hoi & 59.84 & 76.89 & 43.30 & 54.70 & 60.60 & 81.00 & 48.32 & 66.97 \\

        \bottomrule 
    \end{tabular}}
    \caption{Evaluation of WCS and \emph{coref-hoi} on dev sets. NS (No Singletons) refers to annotations with singleton clusters removed. Scores presented are CoNLL F1 scores. Note that the scores are from an internal development set.}
    \label{tab:coref_dev_singletons}
\end{table*}

Using the development sets of the shared task datasets, we evaluated WCS and the \emph{coref-hoi} model. Results are shown in Table \ref{tab:coref_dev_singletons}. Looking at these scores, we found that \emph{coref-hoi} struggled with singleton clusters (as expected), as the CoNLL F1 score of these predictions was much higher after removing the singletons from the annotations. WCS, on the other hand, seemed to do better on singletons than non-singletons, as evidenced by the higher scores on annotations that contain singletons vs. those without. As a result, we combined the strengths of the two systems by simply adding the singletons predictions of WCS to the cluster predictions of \emph{coref-hoi}. This resulted in the highest test set scores (as shown in Table \ref{tab:wcs-combinations}).

\subsection{Training}
\paragraph{WCS}
The WCS system was trained for 5 epochs on Nvidia GeForce RTX 2080. We use teacher forcing for the coreference clusters with a ratio of 30\%. The learning rate is set to 1e-4 and the dropout rate is 0.3. We use Adam as optimizer. It took about 26 hours to train the whole system on the complete training set. 

\paragraph{Coref-HOI}
The \emph{coref-hoi} system was trained for 24 epochs on a Nvidia Quadro RTX 6000. We use a pretrained SpanBERT\textsubscript{Large} model to initialize the base language model. We use a learning rate of 1e-5 for the base model and 3e-4 for the fine tuning layers. We follow all other hyperparameters found in the \textit{train\_spanbert\_large\_ml0\_cm\_fn1000} training configuration of the \emph{coref-hoi} system. Training took about 24 hours.

\subsection{Results and discussion}

Our results on the internal development set as well as on the official test set are reported in Tables ~\ref{tab:wcs-combinations} and \ref{tab:coref_dev_singletons}. Based on the final cluster assignments we can recognize 4 common types of mistakes made by WCS: partial word overlaps (e.g., \textit{`mute button'} and \textit{`volume button'}), embedded mentions (e.g., \textit{`a power supply which we get'} and \textit{`we'}), wrong span boundaries (e.g., \textit{`ok good knight'}) and confusing candidates that have similar surface forms but different meanings (e.g., \textit{`the minutes of uh this meeting'} and \textit{`forty minutes'}). Some of these mistakes were probably caused by the over-reliance of WCS on the head embeddings. Interestingly, when using SpanBERT instead of GloVe and standard BERT for span encoding we observed that many generated clusters contain mentions with spurious connections (e.g., \textit{`the spirits of our people'} and \textit{`such dark superstitions'} or \textit{`the executive'} and \textit{`the company'}).

Judging from the scores on the development set reported in Table \ref{tab:coref_dev_singletons}, WCS shows better performance than \emph{coref-hoi} when the evaluation is done on all clusters including singletons. However, when singletons are excluded \emph{coref-hoi} outperforms WCS and this was the main motivation to combine the outputs of both models. We also evaluated the span extraction performance of WCS vs. the combined system using the gold mention span annotations provided by the shared task organizers. We found that WCS had consistently higher recall but lower precision on mention span detection compared to the combined model. E.g., on the AMI dataset WCS achieved precision 82\% and recall 68\% whereas the combined model achieved precision 84\% and recall 63\%. Similar results were observed on the other two datasets that we tested (Light and Persuasion).

Looking at the mistakes of the combined model we found that some mentions have incorrect spans, e.g., \textit{`half'} and \textit{`hour'} are annotated as two separate mentions in \textit{`see you in half and hour'}. Sometimes the annotated spans are longer than the gold ones, e.g., \textit{`close tabs on you'} instead of \textit{`close tabs'} or \textit{`Of course , good Monk'} instead of \textit{`good Monk'}. This can also result in incorrect clustering such as in case of putting \textit{`this realm'} and \textit{`this realm, stories, population'} in the same cluster. The combined model also struggles with the cases like \textit{`some'} and \textit{`they'} in the following example: \textit{`\textbf{Some} don't give the money out like they are suppose to. Did you heard that \textbf{they} now do every payment taken from people transparent?'} Both mentions were assigned to the same coreference chain although \textit{`some'} should refer to the people who give the money and \textit{`they'} to those who receive it. Despite some problems with the mention span detection the combined model shows overall better clustering performance compared to vanilla WCS.

Experimenting with various combinations of the coreference systems we found that combining the strengths of different systems helps to improve the results. In the future we plan to investigate whether adding coreference signal from the pre-trained models also helps boost the performance and reduce training time for systems like WCS.

For the current submission we combined the model outputs based on some simple heuristics 
but it would be interesting to see whether this process could be also learned by a model. Training a new model from scratch or even fine-tuning it on a new dataset might be sub-optimal or even not feasible in some cases. E.g., when we deal with dialogues instead of narrative texts or if the annotation schemes differ significantly. 
In such cases we believe that using a smart coreference editor that combines and checks outputs of different systems and applies some constraints or filters would be beneficial and we would like to work on such project in the future. 

\section{Discourse Deixis Resolution}
\label{sec:dd-res}


CCST 2022 offers three different tracks for discourse deixis resolution. First track (Eval-DD Pred) assumes finding antecedents for discourse deixis anaphors predicted by models given unannotated data. The second one (Eval-DD Gold M) aims at identification of discourse deixis anaphors among all types of annotated anaphors and non-referential mentions, and their subsequent resolution. The goal of the last track (Eval-DD Gold A) is to find antecedents for already annotated discourse deixis anaphors. Our team participated in all three tracks.

The core of our approach relies on the \emph{coref-hoi} model, because it was successfully adopted for CCST 2021 discourse deixis track by~\citet{kobayashi-etal-2021-neural}. Their model was able to achieve the CoNLL F1 score of 35.4\% - 52.1\% depending on the dataset and shared task track, and ranked first for discourse deixis~\cite{kobayashi-etal-2021-neural}. The summary of our system can be found in Table~\ref{tab:DDR-summary}.

\begin{table*}[h]
    \centering
    \scalebox{0.95}{
    \begin{tabular}{l|l}
        \hline
        \bf{Track} & Resolution of discourse deixis \\
        \hline
        \bf{Setting} & Predicted mentions / Gold mentions / Gold anaphors \\
        \hline
        \bf{Baseline} & The \emph{coref-hoi} model adopted for discourse deixis by~\citet{kobayashi-etal-2021-neural} \\
        \hline
        \bf{Approach} & \shortstack[l]{1) Consider all mentions of \emph{this}, \emph{that}, \emph{it} and \emph{which} potential anaphors \\
        2) Consider all spans in the given segment potential antecedents \\
        3) Represent both anaphor and antecedent candidates as embeddings with additional shallow \\ linguistic features \\
        4) Calculate pairwise anaphor-antecedent scores similar to \emph{coref-hoi} and choose \\ the antecedent based on the largest score \\
        5) Use anaphor-antecedent pair representation to classify the anaphor type and discard \\ non-discourse deictic anaphors} \\
        \hline
        \bf{Train data} & \shortstack[l]{ARRAU corpus (Gnome, Trains\_91, Trains\_93, RST\_DTreeBank, Pear\_stories), AMI, \\ Switchboard, Light and Persuasion} \\
        \hline
        \bf{Dev data} & AMI, Light, Persuasion (dev splits) \\
        \hline
    \end{tabular}}
    \caption{Discourse deixis resolution: approach summary}
    \label{tab:DDR-summary}
\end{table*}

\subsection{Data}

We use training and development data presented in Section~\ref{subsec:ar-data}. 
\emph{Coref-hoi} splits input data into segments of a set length to limit the number of mention candidates. Given a segment, all possible spans/potential mentions are created. Next, this `pool' of mentions is used to form valid anaphor-antecedent pairs. 
In contrast to that, we only consider the occurrences of \emph{`this'}, \emph{`that'}, \emph{`it'} and \emph{`which'} as potential anaphors and treat all other spans in the segment as antecedent candidates. These four markables were chosen based on our observation that they often occur as discourse deixis anaphors in our training data: they make about 72.3\% of all annotated discourse deictic anaphors\footnote{We treat all discourse deictic markables with semantic type `discourse old' as anaphors.}. Similar statistical findings (however, for other dialogue corpora) were reported, e.g., by~\citet{Webber1988DiscourseDA},~\citet{muller2008fully},~\citet{kolhatkar_survey}. Besides being discourse deictic, the markables in focus can also be non-referential (e.g., \emph{`it'} in expletive constructions, \emph{`that'} as a relative pronoun), or anaphoric (e.g., \emph{`this'} as a determiner in a noun phrase).

Because we focus only on certain anaphor candidates, we build segments in a slightly different way than \emph{coref-hoi} does. Instead of splitting the input into non-overlapping chunks of approximately the same length, we go through the input data word by word until any of our anaphors occurs, and then create a segment. Our segment typically includes all (sub)tokens up to the current sentence end to the right of the anaphor, as well as one or more sentences to the left of it. We limit the segment's length by 256 (sub)tokens. Thus, given the same input, we build more segments than \emph{coref-hoi} does, our segments are mostly overlapping, and each one contains only one anaphor candidate.

In total we build 9,827 segments/examples from training data, of which 44\% contain non-referential \emph{`this'}, \emph{`that'}, \emph{`it'} and \emph{`which'}, 41.2\% - anaphoric, and only 14.8\% - discourse deictic ones. To make our training data balanced, we perform undersampling and decrease the number of examples from the first two classes. We end up having 1,454 training samples of each anaphor class. For the sake of simplicity, undersampling is done blindly, i.e. we do not take into consideration how the instances of our three classes are distributed given each of the four markables.

\subsection{Model architecture}


We perform discourse deixis resolution using a multi-task learning approach - besides finding the antecedents, we also need to identify the types of potential anaphors (discourse deictic, anaphoric or non-referential). Type classification is performed after the antecedent (if any) is found. It is also important to emphasize that we 
try to resolve any potential anaphor regardless of its type. Thus, our model also learns to resolve `standard' coreference as a by-product. To our knowledge, our model is the first one doing that.

To perform the resolution, \emph{coref-hoi} first associates each span (represented as an embedding) with a score indicating how likely this span is a valid mention (anaphor or antecedent). To speed up the training process, certain number of spans with the low scores get pruned. Next, the model learns to find the most probable antecedent for each anaphor based on their pairwise scores. 



We modify their approach as follows. First, as we know exactly which span our anaphor $x$ is, and it is the same for all antecedent candidates $y$, we do not score anaphors or calculate pairwise mention scores. An antecedent score $s_m(y)$ is produced by a feedforward neural network $\emph{\text{FFNN}}_m$ taking as input a vector representation of span $y$, like in \emph{coref-hoi}. Second, as shown in Table~\ref{tab:my_mention_rep}, anaphors $k_x=p_x, \rho(x)$ and antecedents $q_y=g_y, \psi(y)$ are composed differently. Main representations $p_x$ and $g_y$ are concatenated with shallow linguistic features $\rho(x)$ and $\psi(y)$ to help our model better differentiate between types of anaphors and antecedent candidates. Our approach to mention representation and motivation behind it are explained in more detail in Section~\ref{subsec:mention-rep}. Third, we do not prune any unlikely antecedents due to the fact that each segment only contains one anaphor, which often has only one antecedent (if mention is anaphoric, there can be more). If we apply pruning, this only antecedent is very likely to be lost at the early stages of training.

\begin{equation} \label{eq2}
\begin{split}
s(x,y) & = s_m(y) + s_f(x,y) + s_s(x,y) \\
s_m(y) & = \emph{\text{FFNN}}_m(q_y) \\
q_y & = g_y, \psi(y) \\
k_x & = p_x, \rho(x) \\
s_f(x,y) & = k_x \cdot q_y \\
s_s(x,y) & = \emph{\text{FFNN}}_c(k_x, q_y, \phi(x,y))
\end{split}
\end{equation}

As shown in Equation group~\ref{eq2}, the final anaphor-antecedent score is the sum of three components: (1) anaphor score $s_m(y)$; (2) fast score $s_f(x,y)$, which is an inner product of vectors $k_x$ and $q_y$ representing anaphor and antecedent, respectively; (3) slow score $s_s(x,y)$, which is an output of a different network $\emph{\text{FFNN}}_c$ taking as input an anaphor-antecedent pair and pairwise features $\phi(x,y)$. Two of pairwise features are borrowed from the \emph{coref-hoi} model. They are distance feature, showing how many sentences/utterances lie between the starting tokens of two mentions, and similarity feature, which is simply a result of am element-wise multiplication of anaphor and antecedent candidate vectors. Finally, we add a token distance feature that shows how many (sub)tokens lie between the starting tokens of the two mentions. This feature is used to help our model learn that in case both anaphor and its antecedent are parts of the same sentence, their starting tokens cannot be close to each other. 

\begin{table}[]
    \centering
    \scalebox{0.75}{
    \begin{tabular}{c|c|c}
        \toprule
        $p_x, \rho(x)$ & $g_y, \psi(y)$ & $\phi(x,y)$ \\
        \midrule
        token emb. & start emb. & sentence dist. emb. \\
        parent emb. & end emb. & token dist. emb. \\
        local context emb. & weighted avg. emb. & similarity emb. \\
        \cmidrule{1-2}
        POS tag emb. & span width emb. &  \\
        DEP tag emb. & span type emb. &  \\
          & end token POS emb. &  \\
          & end token DEP emb. &  \\
        \bottomrule
    \end{tabular}}
    \caption{Representations of anaphor and antecedent candidates, and pairwise features}
    \label{tab:my_mention_rep}
\end{table}

The largest $s(x,y)$ score is used to predict the best antecedent candidate. The antecedent gets concatenated with the anaphor and is used as input for an anaphor type classifier, which is a multilayer perceptron (MLP) network consisting of two linear layers with a ReLU activation function in-between. Similar to \emph{coref-hoi,} to account for the case of non-referential `anaphors', a dummy zero score is always prepended to the row of $s(x,y)$ scores.

\subsection{Mention representation}
\label{subsec:mention-rep}

Potential anaphors and antecedents have different representations. While the main part of an antecedent candidate embedding $g_y$ is constructed similar to \emph{coref-hoi}, the main part of an anaphor embedding $p_x$ is a concatenation of the embedding of the token itself, embedding of the parent token and local context embedding, which includes eight (sub)tokens to the left and right of the anaphor. 

Our decision to include the last two embeddings was motivated by the following observations. Depending on the mention type, mentions' parents have to certain extent different distributions, e.g., discourse deictic mentions more often have forms of the verb \emph{`to be'} as parents than mentions of other two types (see Table~\ref{tab:parent-lemmas-dist} in Appendix~\ref{app:dd}). Moreover, in our data about 60\% of anaphor candidates have verbal parents. And certain verbs (e.g., \emph{`assume'}, \emph{`say'}) are only compatible with discourse deixis~\cite{eckert2000dialogue}. We use SpaCy to identify tokens' parents, and SpanBERT\textsubscript{Large} encoder to acquire tokens' embeddings. The usage of context helps capture various useful patterns that may be characteristic of discourse deixis or identity anaphora. These patterns may include, e.g., adjective-copula constructions. Subjects of such constructions with adjectives applicable to abstract entities (e.g., \emph{`correct'}, \emph{`true'}) usually refer to discourse entities~\cite{eckert2000dialogue}. Other examples are certain types of complement constructions (like \emph{`that is why/because/what/how'}), \emph{`do-object'} expressions,  which also may point at verbal antecedents~\cite{muller2008fully}. The inclusion of context may also be useful for capturing any tokens that point at abstract/concrete character of reference. The size of the context window was chosen intuitively, we did not conduct any separate experiments for finding the optimal window size, but may do it in the future.

Additional linguistic features used to represent anaphors $\rho(x)$ and antecedent candidates $\psi(y)$ are also different. Again, we use SpaCy to extract part of speech (POS) and dependency edge (DEP) tags for tokens in segments, and Berkeley Neural Parser~\cite{kitaev-etal-2019-multilingual} to get syntactic constituents (nominal, verbal, or other). We use POS and DEP tags for anaphors. According to our statistical findings (see Table~\ref{tab:ana-pos-dep-dist} in Appendix\ref{app:dd}), there are some differences in distributions of (POS, DEP) combination depending on the mention type. E.g., the (PRON, nsubj) combination is especially frequent in case of discourse deictic anaphors, while (DET, det) is not. Our antecedent candidates encompass four additional features, of which only span width is borrowed from \emph{coref-hoi}. Other features include span type (verbal, noun, other), POS and DEP tags of the last token. The span type feature was introduced based on the observation that discourse deictic anaphors mostly have verbal phrases or sentences as antecedents, and `standard' anaphors - noun phrases. The other two features are meant to help identify discourse entities, which often encompass the whole sentence and thus end with a punctuation mark. Note that none of our shallow linguistic features is decisive. Moreover, both SpaCy and Berkeley Neural Parser may not function properly on dialogue data. Still, our experiments on the toy dataset (consisting of a single light\_train 2022 file) show that without all these features the model is only able to achieve 29.41\% CoNLL F1 score on the light\_dev 2022 data. Adding features helps increase this score up to 36.44\%. 

All linguistic features described in this section are represented as trainable embeddings of length 100.

\subsection{Training}

To train our model we kept the hyperparameters reported by \emph{coref-hoi}, namely BERT- and task-specific learning rates (1e-5 and 3e-4, respectively), optimizers (AdamW and Adam), schedulers and dropout rate of 0.3. The number of training epochs was set to 24, but we had to stop training after 17 epochs. Currently the model is computationally inefficient (it is able to process only a single training example at a time), so we did not have enough time to complete the training.

The model was trained using a combination of several loss functions: (i) marginal log-likelihood of possibly correct antecedents; (ii) anaphor type loss checking how well the model distinguishes between discourse deixis, identity and non-referential anaphors; (iii) label loss that punishes the model if it tends to reject all antecedent candidates while having a referential anaphor; (iv) constituent type loss checking how well the model can differentiate between valid (verbal and nominal) and invalid (various fragments) antecedents. The addition of label loss is motivated by the fact that at early stages of training our model always tends to reject all antecedents by assigning negative scores to them. Constituent type loss is inspired by the mention loss in \emph{coref-hoi}. The idea is that the model should assign larger scores to valid constituents. This loss is used with a coefficient $\lambda=0.02$ to account for a big number of constituents and prevent it from dominating over all other losses.

\subsection{Results and discussion}

We used the same model for all three discourse deixis tracks. Table~\ref{tab:dd-official-res} illustrates the scores achieved by our model on the official test sets. Because the model is designed to resolve only four potential antecedents, there is no big difference in scores between the (Pred) and (Gold M) tracks. The scores for the latter are even slightly worse, as the model has to deal with numerous anaphor candidates it has not seen before. The best scores are reached for the (Gold A) track. It should be noted that here the model tries to resolve all annotated anaphors, not only the four target ones. Still, we tend to attribute the increase in performance not to a wider coverage of anaphors, but to the fact that the model does not have to classify the anaphor types. 

\begin{table}[]
    \centering
    \scalebox{0.8}{
    \begin{tabular}{lcccc}
        \toprule
        Track & Light & AMI & Persuasion & Swbd. \\
        \midrule
         Eval-DD (Pred) & 36.82 & 50.09 & 47.04 & n/a \\
         Eval-DD (Gold M) & 35.91 & 47.13 & 48.24 & n/a \\
         Eval-DD (Gold A) & 44.95 & 56.54 & 62.79 & n/a \\
        \bottomrule 
    \end{tabular}}
    \caption{CoNLL F1 scores on the official test sets}
    \label{tab:dd-official-res}
\end{table}

\begin{table}[]
    \centering
    \scalebox{0.8}{
    \begin{tabular}{lcccc}
        \toprule
        \multirow{2}{*}{Data} & \multicolumn{2}{c}{2021} & \multicolumn{2}{c}{2022} \\
        \cmidrule{2-5}
         & Our model & Winner & Our model & Winner \\
        \midrule 
        Light & 48.04 & 42.7 & 36.82 & 37.09 \\
        AMI & 40.34 & 35.4 & 50.09 & 53.31 \\
        Persuasion & 56.68 & 39.6 & 47.04 & 54.59 \\
        Swbd. & n/a & 35.4 & n/a & 49.76 \\
        \bottomrule    
    \end{tabular}}
    \caption{Model comparison: CoNLL F1 scores on official tests 2021 and 2022 for the Eval-DD (Pred) track}
    \label{tab:model-comparison}
\end{table}

Table~\ref{tab:model-comparison} shows the CoNLL F1 scores achieved by our system and the winning model on the official test data 2022 for the Eval-DD (Pred) track. Our model ranks second for all the datasets with a score difference ranging from 0.27 to 7.55 points. To compare our model with the baseline model by \citet{kobayashi-etal-2021-neural}, we also evaluate it on the test partitions of Light, AMI and Persuasion datasets without gold annotations released for the CCST 2021. We see that our approach beats the baseline on all the datasets.

To see the limitations of our model and have a better understanding of what it can/cannot learn, we additionally evaluate it on the test partitions of Light, AMI and Persuasion datasets from CCST 2021 containing gold annotations. Our analysis (see Table~\ref{tab:error-analysis} in Appendix~\ref{app:dd}) shows that the model struggles with the anaphor type identification: out of 292 true discourse deictic \emph{`this'}, \emph{`that'}, \emph{`it'} and \emph{`which'} only 212 (72.6\%) are classified as having the same type, 62 (21.25\%) - as anaphoric, and 18 (6.16\%) as non-referential ones. Interestingly, only one of all misclassified anaphors is linked to the correctly predicted antecedent. Also, all anaphors incorrectly classified as non-referential get associated with empty spans. At the same time the model successfully finds antecedents for 144 (67.92\%) out of 212 correctly identified discourse deictic anaphors. It looks like anaphor type is important for the model to be able to perform resolution. 

Looking at Table~\ref{tab:error-analysis}, we can conclude that our model also has difficulties finding split antecedents: 41 anaphors (14.04\%) out of 292 refer to them, but our model only finds 7. 
In general, the model demonstrates a tendency to choose discourse deixis antecedents consisting of single sentences. We hypothesize that it happens for the following reasons. First, there are not enough training examples with split antecedents. Second, our model lacks mechanisms to capture relations between split antecedents making them a coherent piece relative to a discourse deictic anaphor.

The following points should also be emphasized. So far we have not evaluated the performance of our model separately for each of the four anaphor candidates. We have not analyzed the ability of our model to resolve identity anaphora. However, such analysis would be useful, so we plan on conducting it in the future.
Also, using a lot of features slows down the training process. Therefore we are planning to perform experiments testing different combinations of features and various feature embeddings sizes. Additional experiments on how the usage of features influences the model trained on all available training data are also necessary. 
Furthermore, an investigation of the quality of the constituent types, POS and DEP tags would be beneficial, considering that we use SpaCy and Berkeley Neural Parser on dialogue data, while they were trained on text corpora.




\section{Bridging Resolution}
\label{sec:bridging-res}

In this section we introduce our submission for the resolution of bridging references. We submitted to the Eval-Br (Gold A) track, in which gold mentions and anaphors are given. This reduces the problem to the selection of antecedent (from gold mention candidates) for each given anaphor. 

\begin{table}[h]
    \centering
    \scalebox{0.82}{
    \begin{tabular}{l|l}
        \hline
        \bf{Track} & Resolution of bridging \\
        \hline
        \bf{Setting} & Gold mentions and anaphors\\
        \hline
        \bf{Baseline} & \shortstack[l]{Higher order coreference resolution \\ \cite{joshi-etal-2019-bert}} \\
        \hline
        \bf{Approach} & \shortstack[l]{Modify baseline to match setting: \\
            1) Batch size from one document to\\ one anaphor \\
            2) Remove span enumeration step and\\ simple pairwise scorer \\
            3) Use cross entropy loss instead of \\ marginal log-likelihood}\\
        \hline
        \bf{Train data} &  \shortstack[l]{AMI, Switchboard, Light, Persuasion, \\ BASHI, ISNotes} \\
        \hline
        \bf{Dev data} & AMI, Light, Persuasion (dev splits) \\
        \hline
    \end{tabular}}
    \caption{Bridging resolution: approach summary}
    \label{tab:Br-summary}
\end{table}

\subsection{Data}
In addition to the shared task dialogue datasets of AMI (851 bridging instances across 7 documents), Switchboard (603 instances, 11 documents), Light (381 instances, 20 documents), and Persuasion (245 instances, 21 documents), we also utilize the bridging anaphora resolution datasets of BASHI \cite{rosiger-2018-bashi} and ISNotes \cite{isnotes} to train our models. BASHI is a corpus of 50 Wall Street Journal articles, containing 57,709 tokens and 410 bridging pair annotations. ISNotes is a corpus of Wall Street Journal articles as well, containing 663 bridging pair annotations. The inclusion of these supplementary datasets was important, as the shared task datasets are relatively small, and the model architecture is fairly complex and expressive, making it easy to overfit. 

\subsection{Model architecture}
Our approach is based on ``independent'' variant of the higher-order coreference architecture introduced in \citet{joshi-etal-2019-bert}. We make a number of modifications to the architecture and training procedure (an overview of the original model/architecture can be found in \citet{joshi-etal-2019-bert} and the system it is built on, introduced in \citet{lee-etal-2018-higher}. Note that the \emph{coref-hoi} system proposed alternatives to the original higher-order system presented in \citet{joshi-etal-2019-bert}, but these alternatives (such as the cluster merging model variant) are not relevant for our system, as we are not finding clusters of coreferent mentions.

Our modifications follow that of the bridging resolution system introduced in \citet{renner-etal-2021-end}. The first modification is a result of the gold anaphors being given: since we do not need to detect anaphors from the text, we can pass one anaphor at a time into the model (together with the document text and gold mentions) instead of passing the whole document at once and detecting and resolving potential anaphors. While this means potentially processing each document multiple times if there are multiple bridging anaphors in the document, this is done to decrease memory requirements significantly, as the pairwise scoring function is run for just one anaphor with its candidates, instead of many anaphors with all of their candidates. This decrease in memory usage allows for changes to the architecture that make it simpler and more accurate (see next paragraph). Also, in practice, the bridging datasets are relatively small, so this extra processing of the same document results in a negligible decrease in computational efficiency. 

The architecture modifications are made possible by the decrease in memory usage allowed from having the mentions given and processing one anaphor at a time. Recall that in the original architecture by~\cite{lee-etal-2018-higher}, they use a ``two stage beam search'' when detecting mentions and finding coreferent pairs: first, they prune potential mentions based on a span scoring function, then they prune antecedents for each span based on a ``fast'' bilinear scorer (the ``coarse'' part of the coarse-to-fine scorer), before sending the remaining spans and their list of antecedent candidates to the more computation- and memory-heavy ``fine'' scorer. This beam search was proposed to allow the system to scale better to longer documents. By having the gold mentions, we can remove the ``fast'' span scorer from the original model, as we no longer need to enumerate all possible spans. Also, since the pairwise memory restraints are reduced by passing just one anaphor into the model at a time, we can remove the ``coarse'' pairwise scorer and skip directly to the ``fine'' scorer. We make these changes in order to use the more expressive ``fine'' scorer directly on all pairs, without having to filter possible mentions and antecedents based on the less expressive `fast'' span scorer and ``coarse'' pairwise scorer.

After these modifications, the model architecture is as follows: pass entire document through the base contextual language model, obtain span representations for the gold mentions and anaphors, compute antecedents via the higher-order mechanism introduced in \citet{lee-etal-2018-higher}. Also, this allows the use of cross entropy loss over all possible antecedents for each anaphor, instead of the original marginal log-likelihood, leading to a more direct optimization of the pairwise scorer.

We use \texttt{bert-base-uncased} as our base language model. We use this instead of \texttt{bert-large-uncased} because the resulting embedding is of smaller dimensionality, leading to less parameters in our token attention and span pair scoring layers. We experimented with the SpanBERT variant as well, but this led to slightly lower scores in preliminary experiments.

\subsection{Training}
We trained the system for 5 epochs on a single Tesla P100 GPU. The learning rate was set to 3e-3 and we used Adam optimizer. We froze the base BERT model to prevent overfitting as the dataset is relatively small even with the supplementary data, set the dropout to 0.3 in the fine tuning layers, and used a higher-order depth of 2. It took about 1 hour to complete training.

\subsection{Results and discussion}

\begin{table}[]
\centering
\small
\begin{tabular}{cccc}
\toprule
Switchboard & Light & Persuasion & AMI \\ 
\midrule
35.78 & 37.68 & 50.99  & 35.23 \\ 
\bottomrule
\end{tabular}
\caption{Test set results for the bridging task (gold anaphors)}
\label{table:bridging}
\end{table}

The submission Entity-F1 scores are shown in Table \ref{table:bridging}. Overall, we report scores slightly higher than reported in \citet{renner-etal-2021-end} for bridging resolution, with scores on the Persuasion dataset being significantly higher than on the other three datasets. This setting allows for a more direct evaluation of the span embedding and pairwise scoring mechanisms from \citet{joshi-etal-2019-bert} and \citet{lee-etal-2018-higher}, as we can remove steps in the fine tuning architecture that are only needed to manage memory usage. These results show the effectiveness of the span embedding and pairwise score on span comparisons tasks such as gold mention/anaphor bridging resolution.  

\section{Conclusion and Outlook}

In this paper we presented our systems for identity anaphora, bridging and discourse deixis resolution.

Our system for the identity anaphora resolution combines the outputs of WCS and the \emph{coref-hoi} system trained with ``cluster merging''. It ranked second in the shared task competition. When experimenting with WCS we tested different settings and tried replacing and adding different embeddings for mention representations (e.g., SpanBERT). However, the configuration reported in~\citet{anikina-etal-2021-anaphora} turned out to work best on our development set. We also tested a combination of WCS trained on the shared task data and CCS trained on OntoNotes as well as \emph{coref-hoi} trained on a combination of dialogue and non-dialogue datasets. The analysis of the model outputs shows that WCS works reasonably well for detecting singletons and pronominal clusters but performs worse when clustering noun phrases. Hence, we combine the outputs of WCS and the \emph{coref-hoi} model and achieve an average improvement of 7.95\% CoNLL score over vanilla clustering with WCS.

In the future we would like to do a more fine-grained analysis of the combined model outputs and test if one could use automatic coreference annotations from other pre-trained models as a weak supervision signal for WCS. In particular, we are interested in evaluating this model on the domain adaptation task and in the low resource setting. We would also like to perform more experiments with coreference chain editing based on the outputs of several models.

The system for discourse deixis resolution ranked second for all three tracks of the shared task. It was able to reach the CoNLL F1 scores ranging from 35.91\% to 62.79\% depending on the track and dataset. Some of these scores are close to the scores achieved by the winning team. 

The model is based on a novel idea that it is possible to combine the tasks of discourse deixis and anaphora resolution. It is our first attempt at implementing this idea, so there is much space for improvement and additional analysis. First, we plan on making our model computationally more efficient, namely, we are going to perform some experiments with adaptive span pruning and check the influence of linguistic features given a larger training set. Second, it is possible to expand the set of potential anaphors. Before doing that, we need to analyse the ability of our model to resolve identity anaphora. Depending on the results, we may use our discourse deixis resolution model to enhance the coreference resolution performed by the WCS model. Finally, the phenomenon of split antecedents requires more investigation, namely, how we can model coherence/relations between them.

The system for the resolution of gold bridging anaphors is based on a higher order coreference system adapted for the setting. While the gold mentions/anaphors setting is much simpler than full bridging (mention/anaphor detection and resolution), the results show how well the span embedding and pairwise scoring mechanisms from \citet{joshi-etal-2019-bert} and \citet{lee-etal-2018-higher} work for bridging pairs.

\section*{Acknowledgements}

The authors are supported by the German Ministry of Education and Research (BMBF): T. Anikina in project CORA4NLP (grant Nr. 01IW20010); N. Skachkova, P. Trivedi, J. Renner in IMPRESS (grant Nr. 01IS20076). 


\bibliography{anthology,custom}
\bibliographystyle{apalike} 

\clearpage
\appendix

\section{Appendix: Discourse Deixis}
\label{app:dd}

Here we present statistical findings used to pick out features to represent anaphor candidates. Table~\ref{tab:parent-lemmas-dist} shows the relative frequencies of parent tokens' lemmas for three types of `anaphors': discourse deictic, anaphoric and non-referential. Table~\ref{tab:ana-pos-dep-dist} illustrates the joint distribution of POS and DEP labels of possible anaphor candidates, also depending on their type. All numbers were extracted from the CCST 2021 training data, namely ARRAU, Light, AMI, Persuasion, and Switchboard.

\begin{table}[H]
    \centering
    \scalebox{0.75}{
    \begin{tabular}{c|lr|lr|lr}
        \toprule
         & \multicolumn{6}{c}{Anaphor's parent} \\
        \midrule 
         & \multicolumn{2}{c}{DD} & \multicolumn{2}{c}{ID} & \multicolumn{2}{c}{non-ref} \\
        \midrule 
        \parbox[c]{2mm}{\multirow{12}{*}{\rotatebox[origin=c]{90}{Lemma}}} & s & 0.329 & be & 0.139 & be & 0.202 \\
         & be & 0.235 & s & 0.117 & s & 0.078 \\
         & do & 0.040 & have & 0.048 & have & 0.031 \\
         & about & 0.037 & do & 0.038 & like & 0.022 \\
         & sound & 0.020 & use & 0.027 & make & 0.022 \\
         & like & 0.020 & ... & & ... & \\
         & ... & & & & & \\
         & have & 0.017 & & & & \\
         & ... & & & & & \\
         & make & 0.013 & & & & \\
         & ... & & & & & \\
         & use & 0.003 & & & & \\
        \bottomrule 
    \end{tabular}}
    \caption{Distribution of anaphors' parents depending on the anaphors' types}
    \label{tab:parent-lemmas-dist}
\end{table}

\begin{table}[H]
    \centering
    \scalebox{0.6}{
    \begin{tabular}{c|lr|lr|lr}
        \toprule
         & \multicolumn{6}{c}{Mention} \\
        \midrule 
         & \multicolumn{2}{c}{DD} & \multicolumn{2}{c}{ID} & \multicolumn{2}{c}{non-ref} \\
        \midrule 
        \parbox[c]{2mm}{\multirow{5}{*}{\rotatebox[origin=c]{90}{POS+DEP}}} & (PRON, nsubj) & 0.664 & (PRON, nsubj) & 0.46 & (PRON, nsubj) & 0.390 \\
         & (PRON, dobj) & 0.148 & (PRON, dobj) & 0.249 & (SCONJ, mark) & 0.173 \\
         & (PRON, pobj) & 0.117 & (DET, det) & 0.139 & (DET, det) & 0.138 \\
         & (DET, det) & 0.03 & (PRON, pobj) & 0.074 & (PRON, pobj) & 0.110 \\
         & (PRON, mark) & 0.013 & (PRON, dep) & 0.02 & (PRON, dobj) & 0.097 \\
        \bottomrule 
    \end{tabular}}
    \caption{Distribution of anaphors' POS and dependency edges tags depending on the anaphors' types}
    \label{tab:ana-pos-dep-dist}
\end{table}

Table~\ref{tab:error-analysis} presents an error analysis of our discourse deixis resolution model on the test Light, AMI and Persuasion data from CCST 2021. We analyze the antecedent predictions made by our model as follows. If they are not empty, all predicted antecedents are divided into split and not split, depending on a simple heuristics: if a predicted sequence of (sub)tokens (the very last token is always excluded) contains a dot, a question or an exclamation mark), it is considered to be split. Next, we check if the antecedents' borders are correct. Here, four cases are possible: (i) only the left border is wrong; (ii) only the right border is wrong; (iii) both borders are wrong; (iv) both borders are correct. 

The table also shows the anaphor type predicted by the model for all 292 gold discourse deictic anaphors.

\begin{table}[H]
    \centering
    \scalebox{0.75}{
    \begin{tabular}{clcccccc}
        \toprule
         & & \multicolumn{3}{c}{Gold ant. not spl.} & \multicolumn{3}{c}{Gold ant. spl.} \\
         & Predictions & non-ref & DD & ID & non-ref & DD & ID \\
        \midrule
        \parbox[c]{2mm}{\multirow{4}{*}{\rotatebox[origin=c]{90}{not split}}} & left border wr. & 0 & 3 & 3 & 0 & 16 & 0 \\
         & right border wr. & 0 & 1 & 2 & 0 & 0 & 1 \\
         & all borders wr. & 0 & 34 & 45 & 0 & 4 & 7 \\
         & all borders cor. & 0 & 137 & 0 & 0 & 0 & 0 \\
        \midrule 
        \parbox[c]{2mm}{\multirow{4}{*}{\rotatebox[origin=c]{90}{split}}} & left border wr. & 0 & 1 & 0 & 0 & 2 & 0 \\
         & right border wr. & 0 & 3 & 0 & 0 & 0 & 0 \\
         & all borders wr. & 0 & 0 & 0 & 0 & 0 & 0 \\
         & all borders cor. & 0 & 0 & 0 & 0 & 7 & 1 \\
        \midrule
         & empty & 16 & 3 & 3 & 2 & 1 & 0 \\
        \bottomrule
    \end{tabular}}
    \caption{Performance on the test partitions of AMI, Light \& Persuasion datasets from CODI-CRAC 2021 Shared Task}
    \label{tab:error-analysis}
\end{table}

\end{document}